\documentclass[a4paper]{article}
\usepackage{iwslt18,amssymb,amsmath,epsfig,makecell}
\usepackage{hyperref}
\usepackage{xcolor}
\def\reg{{\rm\ooalign{\hfil
     \raise.07ex\hbox{\scriptsize R}\hfil\crcr\mathhexbox20D}}}

\title{ON-TRAC Consortium End-to-End Speech Translation Systems for \\the IWSLT 2019 Shared Task  }

\name{\begin{tabular}{c}Ha Nguyen$^1$, Natalia Tomashenko$^2$, Marcely Zanon Boito$^1$, Antoine Caubri\`ere$^3$, \\
Fethi Bougares$^3$, Mickael Rouvier$^2$, Laurent Besacier$^1$, Yannick Est\`eve$^2$\end{tabular}}

\address{$^1$LIG - Universit\'e Grenoble Alpes, France  \\
$^2$LIA - Avignon Universit\'e, France \\
$^3$LIUM - Le Mans Universit\'e, France \\
}
\begin{document}

\maketitle

\begin{abstract}

This paper describes the ON-TRAC Consortium translation systems developed for the end-to-end model task of IWSLT Evaluation 2019 for the English$\to$ Portuguese language pair.
ON-TRAC Consortium is composed of researchers from three French academic laboratories: LIA (Avignon Universit\'e), LIG (Universit\'e Grenoble Alpes), and LIUM (Le Mans Universit\'e).
A single end-to-end model built as a neural encoder-decoder architecture with attention mechanism was used for two primary submissions corresponding to the two EN-PT evaluations sets: (1) TED (MuST-C) and (2) How2.
In this paper, we notably investigate impact of pooling heterogeneous corpora for training, impact of target tokenization (characters or BPEs), impact of speech input segmentation and we also compare our best end-to-end model (BLEU of 26.91 on   MuST-C  and 43.82 on  How2 validation sets) to a pipeline (ASR+MT) approach.
\end{abstract}


%
\section{Introduction}
\label{sec:intro}

Previous automatic speech-to-text translation~(AST) systems operate in two steps:  source language speech recognition~(ASR) and source-to-target text translation~(MT). However, recent works have attempted to build end-to-end AST without using source language transcription during learning or decoding~\cite{berard-nips2016,weiss2017sequence} or using it at training time only~\cite{berard:hal-01709586}.
Very recently several extensions of these pioneering works were introduced: low-resource AST~\cite{DBLP:journals/corr/abs-1809-01431}, unsupervised AST~\cite{DBLP:journals/corr/abs-1811-01307}, end-to-end speech-to-speech translation~(\textit{Translatotron})~\cite{DBLP:journals/corr/abs-1904-06037}. Improvements of end-to-end AST were also proposed using weakly supervised data~\cite{DBLP:journals/corr/abs-1811-02050}, or by adding a second attention mechanism~\cite{DBLP:journals/corr/abs-1904-07209}.

This paper describes the ON-TRAC consortium automatic speech translation~(AST) systems for the IWSLT 2019 Shared Task. ON-TRAC Consortium is composed of researchers from three French academic laboratories: LIA~(Avignon Universit\'e), LIG~(Universit\'e Grenoble Alpes), and LIUM~(Le Mans Universit\'e).

We participated to the end-to-end model English-to-Portuguese AST task on \textit{How2} \cite{sanabria2018how2} and \textit{MuST-C} \cite{di2019must} datasets. We notably try to answer to the following questions:
\begin{itemize}
    \item Question~1: does pooling heterogenous corpora (\textit {How2} and \textit{MuST-C}) help the AST training? 
    \item Question~2: what is the better tokenization unit on the target side (BPE or characters)?
    \item Question~3: 
    considering that segmentation is an important challenge of AST, what is the optimal way to segment the speech input?
    \item Question 4: 
    does fine-tuning increase the system's performance?
    \item Question 5: is our end-to-end AST model better than an ASR+MT pipeline?
    
\end{itemize}
    
This paper is organized as follows: after briefly presenting the data in Section~\ref{sec:data} and after detailing our investigation on automatic speech segmentation in Section~\ref{sec:seg}, we present the end-to-end speech translation systems submitted by our ON-TRAC consortium in Section~\ref{sec:systems}. Section~\ref{sec:lessons} summarizes what we learned from this evaluation and Section~\ref{sec:conclusion} concludes this work.




\section{Data}
\label{sec:data}


The corpora 
used in this work are the How2~\cite{sanabria2018how2} and MuST-C~\cite{di2019must} corpora. Since 
we focus on English-to-Portuguese AST tasks, only the English-Portuguese portion of MuST-C corpus is used. 
The statistics of these two corpora, along with the corresponding provided evaluation data, can be found in Table~\ref{table:1}. 
In order to answer to the first scientific question mentioned in Section~\ref{sec:intro}, we pool these two corpora together to create a merged corpus whose details can also be found in the same table.

\begin{table}[h!]
 \centering
 \begin{tabular}{| l | c | c | c | c |} 
  \hline
  \textbf{\thead{Corpus}} & \textbf{\thead{\#Segments}} & \textbf{\thead{Hours}} & \textbf{\thead{\#src \\ words}} & \textbf{\thead{\#tgt \\ words}}\\ [0.5ex] 
  \hline
  \hline
  MuST-C & 206,155 & 376.8 & 3.9M & 3.7M \\ 
  \hline
  How2 & 184,624 & 297.6 & 3.3M & 3.1M \\
  \hline
  Merged corpus & 390,779 & 674.4 & 7.2M & 6.8M \\
  \hline
  \hline
  MuST-C eval & 2,571 & 5.4 & - & - \\ 
  \hline
  How2 eval & 2,497 & 4.5 & - & - \\
  \hline
 \end{tabular}
 \caption{Statistics of the original MuST-C and How2 corpora, the merged version, and the official evaluation data (audio data only).}
 \label{table:1}
\end{table}

Note that the statistics for the How2 training set might slightly differ 
from that of other participants since 
the original audio files for the How2 corpus are not officially available.
Since we wanted to apply our own feature extraction, instead of using the one shared by the How2 authors, 
we have to download the original video files from Youtube,\footnote{https://www.youtube.com/} and then extract the 
audio from these downloaded video files. 
One issue with 
this approach is that the final corpus content will depend on 
the availability of audio files on Youtube 
at the downloading date. On July 12th, when our version of the corpus was downloaded, 21 (out of 13,472) video files were missing. We consider this as a minor loss with regard to the possibility it gives us to extract our own acoustic features.

\section{Speech segmentation}
\label{sec:seg}

While How2 evaluation data is distributed with a predefined segmentation, this information is not provided for the TED talks evaluation data. In this context, we explore two different approaches to segment the MuST-C (TED talks) audio stream.
The first one is based on the use of the well known LIUM\_SpkDiarization toolkit~\cite{meignier2010lium}, which is an open source toolkit for speaker diarization (we used the default configuration).

The second approach is based on the use of an Automatic Speech Recognition system~(ASR) as a speech segmenter: we transcribe automatically and without segmentation all the validation and evaluation datasets with a Kaldi-based ASR system~\cite{daniel2011kaldi} trained on  TEDLIUM~3~\cite{hernandez2018ted}.\footnote{In the context of the campaign, the use of some of 
TEDLIUM~3 files is forbidden. These files have been removed before training the ASR system.} We did not try to optimize the ASR system on our data.
This ASR system produces recognized words with timecodes (start time and duration for each word). Thanks to this temporal information, we are able to measure silence duration between two words when silence (or non speech event) exists. When a silence between two words is higher than 0.65 seconds, we split the audio file. When the number of words in the current speech segment
exceeds 40, this threshold is reduced to 0.15 seconds, in order to avoid exceedingly
long segments. These thresholds have been tuned in order to get a segment duration distribution in the evaluation data close to the one observed in the training data.
Table~\ref{table:stattrainseg} summarizes statistics about segment duration on training data (with the segmentation provided by the organizers) and evaluation data (ASR-based segmentation \textsl{vs.} speaker diarization toolkit).

\begin{table}[h!]
 \centering
 \begin{tabular}{| l | c | c | c | c |} 
  \hline
  \textbf{\thead{Corpus/Segmentation}} & \textbf{\thead{min size}} &  \textbf{\thead{max size}} & \textbf{\thead{average}} & \textbf{\thead{std dev}}\\ 
  \hline
  \hline
  Train/Organizers & 0.17 & 30.00 & 6.31 & 4.72  \\ 
  \hline
  \hline
  Eval/ASR-based  & 0.03 & 22.71 & 6.09 & 4.52   \\ 
  \hline
  Eval/SpkDiar & 1.51 & 20.00 & 9.62 & 5.33 \\
  \hline
 \end{tabular}
 \caption{Statistics on speech segments duration (MuST-C) for 2 different segmentation approaches. All values are given in seconds.}
 \label{table:stattrainseg}
\end{table}

In order to choose the segmentation process for our primary system among these two approaches, we carried out experiments on the tst-COMMON data from the MuST-C corpus.
For these experiments, we applied a preliminary version of our end-to-end system, trained on the MuST-C training data to translate speech into lower-case text. 
Then, we used the \textit{mwerSegmenter} tool\footnote{\url{https://www-i6.informatik.rwth-aachen.de/web/Software/mwerSegmenter.tar.gz}} to realign our translations to the reference segmentation of the tst-COMMON data, in order to evaluate translation quality. 
Table~\ref{table:mWER+BLEUseg} shows the BLEU score obtained with different segmentation strategies: manual (original MuST-C annotations), ASR-based, and speaker diarization.

\begin{table}[h!]
 \centering
 \begin{tabular}{| l | c | } 
  \hline
  \textbf{\thead{Segmentation}} & \textbf{\thead{BLEU}}\\ 
  \hline
  \hline
  Manual (original) & 25.50  \\ 
  \hline
  \hline
  Speaker Diarization \cite{meignier2010lium}  & 21.01   \\ 
  \hline
  ASR-based & 22.03 \\
  \hline
 \end{tabular}
 \caption{BLEU scores (lower-case evaluation) obtained on the tst-COMMON (MuST-C corpus) data with different speech segmentation strategies.}
 \label{table:mWER+BLEUseg}
\end{table}

Those preliminary results show that ASR-based segmentation leads to better speech translation performance than the speaker diarization approach. However, we observe that manual segmentation~(25.50) still outperforms 
our best automatic segmentation~(22.03). This shows that automatic segmentation of the audio stream is an important issue to address for the speech translation task.

Finally, 
based on these findings we decided to use the ASR-based approach for our primary system applied to the TED talks (MuST-C) evaluation data (for which we do not possess manual segmentation). For the How2 evaluation data, we use the manual segmentation provided by the organizers.



\section{Speech translation systems }
\label{sec:systems}


In this work, several 
speech translation systems were developed for translating English speech into Portuguese text (EN-PT). 

\subsection{End-to-end speech translation}

In this section we detail our end-to-end architecture. All the experiments presented 
are conducted using the ESPnet~\cite{watanabe2018espnet} end-to-end speech processing toolkit.

\textbf{Speech features.}
For all models, 80-dimensional Mel filter-bank features, concatenated with 3-dimensional pitch features,\footnote{Pitch-features are computed using the Kaldi toolkit~\cite{daniel2011kaldi} and consist of the following
values~\cite{ghahremani2014pitch}: (1) probability of voicing (POV-feature), (2) pitch-feature and (3) delta-pitch feature. For details, see \url{http://kaldi-asr.org/doc/process-kaldi-pitch-feats_8cc.html} }
are used for training. 
Features are extracted using $25ms$ windows with a frame shift of $10ms$. Cepstral mean and variance normalization is computed on the training set. 
Data augmentation, based on speed perturbation with factors of 0.9, 1.0, and 1.1, is applied to the training data~\cite{ko2015audio}.

\textbf{Text preprocessing.} 
Following the ESPnet speech translation recipe, we normalize punctuation, and tokenize all the Portuguese text using Moses.\footnote{\url{http://www.statmt.org/moses/}}
Texts are case-sensitive and contain punctuation.
Moreover, the texts of the MuST-C corpus contain \textit{'Laughter'}, \textit{'Applause'} marks. These are kept for 
training 
the model which uses only 
MuST-C data, 
but they are removed from the texts
when training the models 
on the combination of both corpora to ensure consistency.

The development sets are generated by randomly sampling 2,000, 2,000, and 4,000 sentences from MuST-C, How2 and the merged corpus respectively. These sentences are 
removed from the corresponding training sets.

Furthermore, to make the training feasible with our limited computational resources, training and development sentences longer than $3,000$ frames ($\approx30s$) or $400$ characters are removed. This results in  $6\%$, $8\%$ and $7\%$ speech data loss for How2, MuST-C and the merged corpus respectively.

The summarization of the training data after preprocessing can be found in Table~\ref{table:preprocessed_data}.

\begin{table}[h!]
 \centering
 \begin{tabular}{| l | c | c | c | c |} 
  \hline
  \textbf{\thead{Set}} & \textbf{\thead{\#Segments}} & \textbf{\thead{\#src \\ words}} & \textbf{\thead{\#tgt \\ words}}\\ [0.5ex] 
  \hline
  \hline
  MuST-C train & 597,871 & 10.9M & 10.3M \\ 
  \hline
  MuST-C dev & 1,994 & 36.4K & 34.4K \\
  \hline
  \hline
  How2 train & 538,231 & 9.4M & 8.9M \\
  \hline
  How2 dev & 1,984 & 33.7K & 32.0K \\
  \hline
  \hline
  Merged train & 1,136,084 & 20.9M & 19.2M \\
  \hline
  Merged dev & 3,978 & 72.4K & 66.5K \\
  \hline
 \end{tabular}
 \caption{Statistics for the training data after preprocessing.}
 \label{table:preprocessed_data}
\end{table}

\begin{figure}
    \centering
    \includegraphics[scale=0.4]{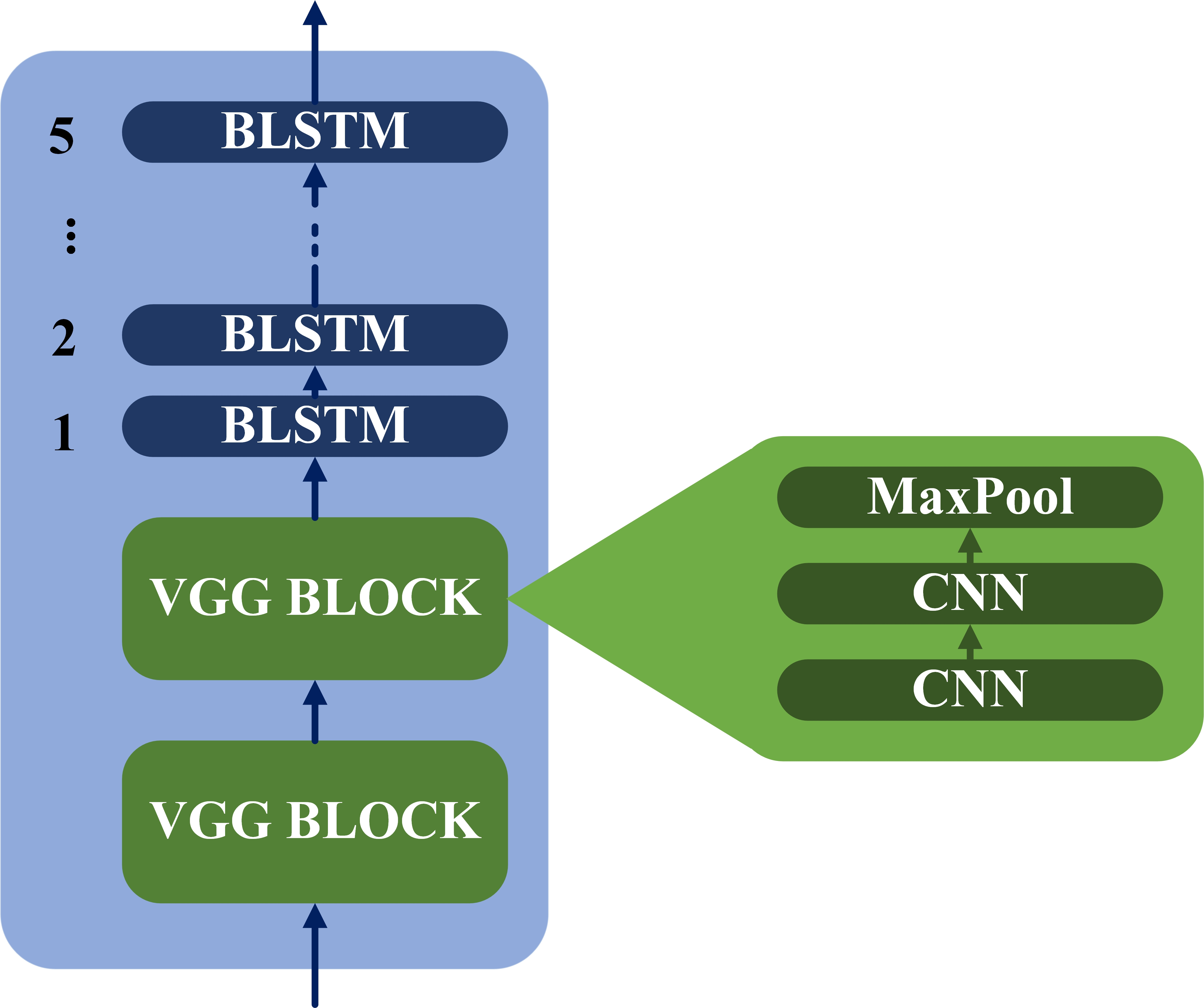}
    \centering
    \caption{Architecture of the speech encoder: a stack of two VGG blocks followed by 5 BLSTM layers.}
    \label{fig:1}
\end{figure}

\textbf{Architecture.}
We use an attention-based encoder-decoder architecture, whose  
encoder has two VGG-like~\cite{simonyan2014very} CNN blocks 
followed by five stacked 1024-dimensional  BLSTM layers (see Figure~\ref{fig:1}). The decoder has two 1024-dimensional LSTM layers.
Each VGG block contains two 2D-convolution layers followed by a 2D-maxpooling layer whose aim is to reduce both time ($T$) and frequency dimension ($D$) of the input speech features by a factor of $2$. These two VGG blocks transform input speech features' shape from $(T \times D)$ to $(T/4 \times D/4)$. 
Bahdanau's attention mechanism~\cite{bahdanau2014neural} is used in all our experiments.


\textbf{Hyperparameters' details.}
In all our experiments, dropout is set only on the encoder part with the probability of $0.3$. Adadelta is chosen as our optimizer. 
%
All  end-to-end models developed in this paper have similar architectures and differ mainly in the following aspects:
(1) training corpus; (2) type of tokenization units;  (3) fine-tuning and pretraining strategies. 
Description of different models and evaluation results are given in Section~\ref{sec:lessons}.


\subsection{Pipeline approach (baseline)}

In this section we describe the pipeline approach for speech translation. 

\textbf{ASR system.}  Kaldi speech recognition tookit \cite{Povey11thekaldi} was used for this purpose. The system used in the pipeline is close to the \textit{tedlium/s5\_r3} recipe.\footnote{\url{https://github.com/kaldi-asr/kaldi/tree/master/egs/tedlium/s5_r3/}} The acoustic model is trained on TEDLIUM-3 and a subset of MuST-C corpus. We use TDNN-F (11 TDNN-F layers) structures for acoustic modeling with 40-dimensional MFCC features. A simple 3-gram language model (LM) is trained using TEDLIUM-3, MuST-C and How2 corpus, with SRILM toolkit \cite{Stolcke02srilm}. 
The ASR system achieved a case-insensitive Word Error Rate (WER) of  21.71\% and 26.89\% on \emph{Must-C tst-COMMON} and \emph{How2 val} sets respectively.


\textbf{MT system.} we used the Transformer~\cite{NIPS2017_7181} sequence-to-sequence model as implemented  in \texttt{fairseq}~\cite{ott2019fairseq}. 
Transformer is the state of the art NMT model. In this architecture, scaled-dot-product attention between keys, values and query vectors in multiple dimensions (or heads) is computed. This is done both \textit{within} encoder and decoder stacks (multi-head self attention) and \textit{between} encoder and decoder stacks (multi-head encoder-decoder attention). 

Our models are based on the small transformer settings using 6 stacks (layers) for encoder and decoder networks with an embedding layer of size 512, a feed-forward layer with an inner dimension of 1024, and 4 heads for the multi-head attention layers. We 
train the NMT system using the merged corpora (Table \ref{table:1}) with a vocabulary of 30K units based on a joint source and target byte pair encoding (BPE) \cite{sennrich2016neural}. Results of the pipeline speech translation system are 
reported in the last line of Table~\ref{table:pipeline}.
\begin{table}[h!]
 \centering
 \begin{tabular}{| l | c | c |} 
  \hline
  \textbf{\thead{Evaluation set}} & \textbf{\thead{ASR }} & \textbf{\thead{Ref }} \\
  \hline
  \hline
  How2 val & 34.23 & 51.37   \\ \hline
  MuST-C tst-COMMON  & 22.14  &28.34    \\\hline
  
 \end{tabular}
 \caption{Detokenized Case-sensitive BLEU scores  for  different evaluation sets when translating the automatic (ASR) and human (Ref) transcription. }
 \label{table:pipeline}
\end{table}


\section{Experiments and lessons learned}
\label{sec:lessons}
In order to answer the scientific questions introduced in Section~\ref{sec:intro}, we conducted a series of 
experiments whose results are presented in Table~\ref{table:2}. 

\subsection{Question 1: choosing the training corpus}
We train three end-to-end models with the architecture described in Section~\ref{sec:systems} using three different training corpora:
(1)  MuST-C, (2) How2, and (3) the merged version of the two corpora. The target tokens are characters. 
These models are then evaluated on the \textit{tst-COMMON} (MuST-C), and \textit{val} (How2) datasets, and the results  are reported in the first three lines of  Table~\ref{table:2}. We can observe that 
the model trained on the merged corpora outperforms the ones trained on MuST-C (difference of 3.32) and How2 (difference of 3.11). 
This model (line \#3 of the table) is used for our IWSLT primary system submission for both evaluation datasets.

\subsection{Question 2: choosing the tokenization units}

In this series of experiments, we investigate the impact of the tokenization units on the performance of the translation system.
We investigated two types of tokenization units: characters and subword units based on byte-pair encoding~(BPE)~\cite{sennrich2016neural}.
Using BPE units, we train four models with different vocabulary sizes: 
400, 2,000, 5,000 and 8,000. 
Results for the 
models are given in Table~\ref{table:2}, lines \#4--7, in which we observe that having fewer output tokens on the decoder side is beneficial. We conclude that characters seem to be the best tokenization units on the MuST-C, and BPE 400 units provides the best results for the How2 task.\footnote{However, since the bpe-400 result for How2 was obtained after the evaluation deadline, our official submission uses characters for both 
datasets).}

\begin{table}[h!]
 \centering
 \begin{tabular}{| c | l | c | c | c |} 
  \hline
  \textbf{\thead{No.}} & \textbf{\thead{Experiment}} & \textbf{\thead{Token}} & \textbf{\thead{MuST-C \\ tst-COMMON}} & \textbf{\thead{How2 \\ val}} \\ 
  \hline
  \hline
  1 & Must-C & char & 23.59 & - \\ 
  \hline
  2 & How2 & char & - & 39.86 \\
  \hline
  3* & Merged & char & \textbf{26.91} & 42.97 \\
  \hline
  4 & Merged & bpe-400 & 24.73 &  \textbf{43.82} \\
  \hline
  5 &  Merged &   bpe-2k &   {23.11}  &  {41.45} \\
  \hline
  6 & Merged & bpe-5k & 22.25 & 41.20 \\
  \hline
  7 & Merged & bpe-8k & 21.75 & 40.07 \\
  \hline
  8 & FT / Unfreeze & char & - & 43.02 \\
  \hline
  9 & FT / Freeze & char & - & 43.04 \\
  \hline
  \hline
  10 & Pipeline (table \ref{table:pipeline})  & bpe-30k & 22.14 & 34.23 \\
  \hline
 \end{tabular}
 \caption{Detokenized case-sensitive BLEU scores  for  different experiments. Two lines with $FT$  correspond to the models trained on the merged training corpus and fine-tuned (FT) using only the How2 corpus. }
 \label{table:2}
\end{table}

\subsection{Question 3: segmentation}

We have seen in Section~\ref{sec:seg} that our ASR-based segmentation leads to better BLEU scores than using off-the-shelf speaker diarization. Our primary system used the ASR-based segmentation to process TED talks, while a contrastive system used  speaker diarization. We expect that the final campaign results will confirm our preliminary conclusion.\footnote{This was written before the evaluation campaign final results release.}

\subsection{Question 4: fine-tuning impact}

We also investigate fine-tuning. For instance, 
training for one more epoch on the target corpus might help to improve translation performance. In order to verify this, we extend the training of the model which uses the merged corpora (line \#3 in Table~\ref{table:2}) for one more epoch on the How2 corpus only (our evaluation target). We investigated (1) fine-tuning both encoder and decoder (\textit{Unfreeze}, line \#8) and (2) fine-tuning  the decoder only (\textit{Freeze}, line \#9). 
Results are presented at the last two lines of Table~\ref{table:2}. We observe a slight but not significant gain with fine-tuning and no difference between \textit{Freeze} and \textit{Unfreeze} options.

\subsection{Question 5: pipeline or end-to-end}
The pipeline results for both corpora are available in the last line (\#10) of Table~\ref{table:2}. We verify that our best end-to-end speech translation results (lines \#3 and \#4) outperform this baseline model by a difference of 4.77 points for TED talks and 9.59 points for How2. While it is important to mention that we did not fully optimize ASR, NMT systems and their  combination,\footnote{ASR and MT were developed independently of each other in two different research groups}
we find that these results highlight the performance of our end-to-end speech translation systems.

\section{Conclusion}
\label{sec:conclusion}
This paper  described the ON-TRAC consortium submission to the end-to-end speech translation systems for the IWSLT 2019 shared task. Our primary end-to-end  translation  model used in the IWSLT-2019,  evaluated  on  the  development  datasets,  scores the following results for case-sensitive BLEU score: 26.91 on TED talks task and 43.02 on How2. 
For the How2 task, we verified (after the evaluation campaign deadline) that it is possible to obtain a better result by using the model with 400 BPE units.

\section{Acknowledgements}

This work was funded by the French Research Agency (ANR) through the ON-TRAC project under contract number ANR-18-CE23-0021.

\bibliographystyle{IEEEtran}

\bibliography{refs}
\end{document}